# Thrill-K Architecture: Towards a Solution to the Problem of Knowledge Based Understanding

Gadi Singer[1], Joscha Bach[1], Tetiana Grinberg*[1][0000-0003-1179-5323], Nagib Hakim[1], Phillip R Howard[1], Vasudev Lal[1] and Zev Rivlin[1]

[1] Intel Labs, Santa Clara, CA 95054, USA
`{gadi.singer, joscha.bach, tetiana.grinberg, nagib.hakim, phillip.r.howard, vasudev.lal, zev.rivlin}@intel.com`

**Abstract.** While end-to-end learning systems are rapidly gaining capabilities and popularity, the increasing computational demands for deploying such systems, along with a lack of flexibility, adaptability, explainability, reasoning and verification capabilities, require new types of architectures. Here we introduce a classification of hybrid systems which, based on an analysis of human knowledge and intelligence, combines neural learning with various types of knowledge and knowledge sources. We present the Thrill-K architecture as a prototypical solution for integrating instantaneous knowledge, standby knowledge and external knowledge sources in a framework capable of inference, learning and intelligent control.

## 1 Introduction: The Rise of Cognitive AI

Many of the current Deep Learning (DL) applications address perception tasks related to object recognition, natural language processing (NLP), translation, and other tasks that involve broad data correlation processing such as that performed by recommendation systems. DL systems yield exceptional results based on differential programming and sophisticated data-based correlation and are expected to drive transformation across industries for years to come. At the same time, a number of fundamental limitations inherent to the nature of DL itself must be overcome so that machine learning, or more broadly AI, can come closer to realizing its potential. A concerted effort in the following three areas is needed to achieve non-incremental innovation:

- Materially improve model efficiency (e.g., reduce the number of parameters by two to three orders of magnitude without loss in accuracy)
- Substantially enhance model robustness, extensibility, and scaling
- Categorically increase machine cognition

Among other developments, the creation of transformers and their application in language modeling [1] has driven computational requirements to double roughly every 3.5 months in recent years [2], highlighting the urgency for improvements in model efficiency. Despite developments in acceleration and optimization of neural networks, without improvements in model efficiency, current model growth trends will not be sustainable for the long haul [3].

Techniques such as pruning, sparsity, compression, distillation and graph neural networks (GNNs) offer helpful advancements in efficiency but ultimately yield only incremental or task specific improvements. Reducing model size by orders of magnitude without compromising the quality of results will likely require a more fundamental change in the methods for capturing and representing information itself and in the learning capabilities within a DL model. Using AI systems that integrate neural networks with added information injected per-need might help to scale down some of the language model growth trends. On a more fundamental level, deep learning lacks the cognitive mechanisms to address tasks central to human intelligence, missing competencies such as abstraction, context, causality, explainability, and intelligible reasoning.

There is a strong push for AI to reach into the realm of human-like understanding. Leaning on the paradigm defined by Daniel Kahneman in his book, Thinking, Fast and Slow [4], Yoshua Bengio equates the capabilities of contemporary DL with what Kahneman characterizes as "System 1" capabilities — intuitive, fast, unconscious, habitual thinking [5]. In contrast, he posits that the next challenge for AI systems lies in implementing the capabilities of "System 2" — slow, logical, sequential, conscious, and algorithmic thinking, such as those needed in planning and reasoning. In a similar fashion, Francois Chollet describes [6] an emergent new phase in the progression of AI capabilities based on broad generalization ("Flexible AI"), capable of adaptation to unknown unknowns within a broad domain. Both these characterizations align with DARPA's Third Wave of AI [7], characterized by contextual adaptation, abstraction, reasoning, and explainability, and by systems constructing contextual explanatory models for classes of real-world phenomena. One possible path to achieving these competencies is through the integration of DL with symbolic reasoning and deep knowledge. We use the term Cognitive AI to refer to this new phase of AI.

There is a divide in the field of AI between those who believe categorically higher machine intelligence can be achieved solely by advancing DL further, and those who do not. Taking the neuro-symbolic side of this debate, we see the need for incorporating additional fundamental mechanisms while continuing to advance DL as a core capability within a larger architecture. Knowledge that is structured, explicit, and intelligible can provide a path to higher machine intelligence or System 2 type capabilities. Structured knowledge is required to capture and represent the full richness associated with human intelligence, and therefore constitutes a key ingredient for higher intelligence. Such knowledge enables abstraction, generalization to new contexts, integration of human generated expertise, imagination of novel situations, counterfactual reasoning, communication and collaboration, and a higher degree of autonomous behavior. If developed, Cognitive AI will be characterized not only by the ability to access and represent symbolic knowledge in conjunction with learning mechanisms, but also by the ability to integrate this knowledge, use it for reasoning, planning, decision making and control, and generate new knowledge via inference and abstraction.

## 2   Dimensions of Knowledge

What we call "human knowledge" encompasses a diverse set of models and information types. We introduce here a classification of the different dimensions of human knowledge, from which we can extrapolate to machine knowledge and understanding.

We distinguish between six dimensions of knowledge. Three are dimensions of direct knowledge, two are meta-dimensions (context and values), and one allows for connecting references (ConceptRefs) (Fig. 1):

1. **Descriptive knowledge** consists of conceptual abstractions and can also include facts and systems of records. The facts and information relevant for specific use cases and environments can be organized, utilized and updated as hierarchical knowledge. The underlying ontology used in individual AI systems can be seeded with task-relevant classes and entities from curated systems (e.g., the OpenCyc ontology or the AMR named entity types).
2. **Dynamic models** of the world include physical, mathematical/algebraic, financial, perceptual and other structural regularities and abstractions that describe how the observed environment will likely change given its current state. Dynamic models can be formal, but when dealing with the heterogeneity of the real world, they often merely capture statistical regularities. Causal knowledge enables going beyond statistical prediction by identifying the conditions under which events manifest, which is a prerequisite for planning and explainability.
3. Humans often use **stories** and **scripts** to provide frames and contexts for the interpretation of facts and events. Stories can take the form of complex narratives that build on shared beliefs and mythologies.
4. **Context and source attribution** is a meta-knowledge dimension that enables the binding of knowledge to a particular context, dealing with conflicting knowledge, and retracting and updating it when other sources are available. Source attribution can become particularly relevant when a system has to deal with questions of data provenance and information reliability (e.g. from news sources).
5. **Values and priorities** are meta-knowledge dimensions that enable specifying the relevance of knowledge and the contexts in which systems can choose a course of action over another so as to behave according to ethical considerations and normative constraints.
6. **Concept references** enable binding different types, modalities and instances of representations together, and unifying knowledge by identifying the relationships between the representations of unique entity. A Concept Reference (or ConceptRef for short) is the identifier and set of references to all things related to a given concept.

The ConceptRefs themselves do not actually include any of the knowledge — the knowledge resides in the dimensions described above.

Understanding requires a foundation of **common-sense knowledge**: a broad (and broadly shared) set of unwritten assumptions that humans automatically apply to make sense of the world. In our framework, commonsense knowledge is considered a subset of each of the above five knowledge dimensions.

For AI systems, implementing knowledge dimensions observed in human comprehension and communication can provide substantial value to the system's intelligence, constituting what we term deeply structured knowledge.

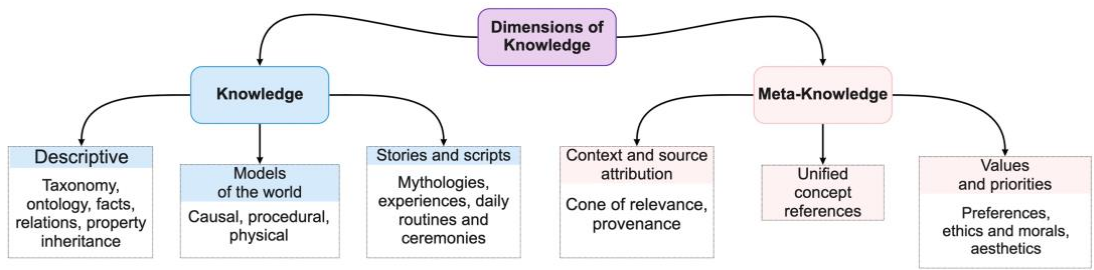

*Figure 1. Dimensions of knowledge*

## 3 From Knowledge to Understanding

Now that we have introduced a taxonomy of human knowledge, we can build upon it to create a definition of understanding that can be reasonably applied to AI systems.

Understanding is the foundation of intelligence. Yoshua Bengio characterizes human-level AI understanding as follows [8]: capture causality and how the world works; understand abstract actions and how to use them to control, reason and plan, even in novel scenarios; explain what happened (inference, credit assignment); and generate out-of-distribution.

We are proposing a knowledge-centric definition of understanding: the ability to create a persistent world view expressed in rich knowledge representation; the ability to acquire and interpret new information to enhance this world view; and the ability to effectively reason, decide and explain existing knowledge and new information.

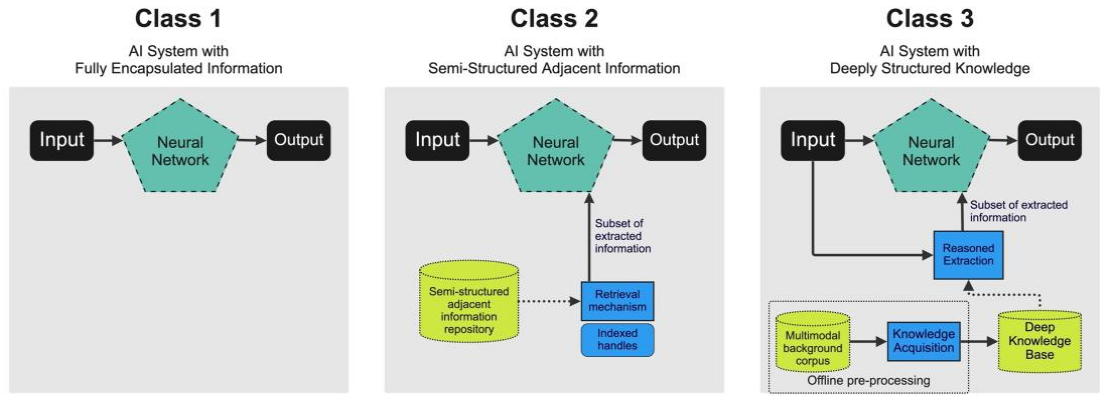

*Figure 2. Information-centric classification of AI systems*

## 4 Information-Centric Classification of AI Systems

The above knowledge-centric definition of understanding leads us to an information-centric classification of AI systems (Fig. 2) as a complementary view to a processing-based classification such as Henry Kautz's taxonomy for neuro-symbolic computing [9]. The classification emphasizes the high-level architectural choice related to the structure and use of information in the AI system.

The proposed information-centric classification includes three key classes of AI systems based on the architectural partition and the use of information on the fly during inference time:

- **Class 1 — Fully Encapsulated Information:** Training data and relations are incorporated into the parametric memory of the neural network (NN) itself. There is no access to additional information at inference time. Examples include recent end-to-end deep learning (DL) systems and language models (e.g., GPT-3). Such systems will likely be the best solutions for all types of perception tasks (such as image recognition and segmentation, speech recognition, and many natural language processing functions), sequence-to-sequence capabilities (such as language translation), recommendation systems, and various question-and-answer applications.
- **Class 2 — Semi-Structured Adjacent Information** (in retrieval-based systems): These systems rely on retrieving information from an external repository (e.g., Wikipedia) in addition to the NN parametric memory (e.g., retrieval-augmented generation) [10]. In Class 2 systems, the repository contains information, but much of the complex relationships and insights related to the information are encapsulated in the embedded space of the NN. These systems are most helpful in addressing use cases with very large data/information space, e.g., an AI system tasked with answering questions about Wikipedia articles. The ability to modify the information in the repository between training time and inference time can be important for out-of-domain challenges, even if the relevant information was not present during training.

- **Class 3 — Deeply Structured Knowledge** (in retrieval-based systems): Retrieval-based systems that interact closely with a deeply structured knowledge base, where the latter uses an explicit information structure that incorporates the multiple knowledge dimensions and their complex relations described in Dimensions of Knowledge (Section 2). The major distinction between Class 2 and Class 3 AI systems lies in where the deeper knowledge resides — whether in the NN parametric memory (Class 2) or in the form of deeply structured knowledge in a knowledge base (knowledge graph in Class 3) tightly integrated with the NN. Class 3 systems could create a multi-faceted reflection of the outside world within the AI system by growing their deep knowledge base through interaction with the training data and classifying the concepts they accrue using the humanlike dimensions of knowledge described above.

### 4.1 Key Elements of a Class 3 System

In an AI system with deeply structured knowledge (Class 3), an NN has an adjacent knowledge base with an explicit structure that conveys the relations and dependencies that constitute deep knowledge. The auxiliary knowledge base is accessed both during training and inference time. Some of the deep knowledge still resides in the NN parametric memory, but in this class of systems, most of the knowledge resides outside the NN. In an NN-only reasoning system, the knowledge base serves as a repository. A Class 3 system will use an explicit knowledge base during inference; however, reasoning functions such as sorting, selection, neighbor identification, and others are conducted by the NN within the embedded space — as can be found in examples of QA systems operating over knowledge graphs [11].

Other Class 3 systems could have an active functionality for selecting information or performing parts of the reasoning on top of the Knowledge Base (KB). We refer to such a mechanism as **reasoned extraction**. One example is Neuro-Symbolic Question Answering (NSQA) [12]. A key advantage of reasoned extraction over NN-only reasoning is that the answer returned by the system can change dynamically as the Knowledge Graph (KG) is updated, without needing to retrain the model.

The rightmost section of Fig. 2 depicts the high-level architecture of a Class 3 system and its key components. The term **Knowledge** refers to the relevant and objective information gained through experience. **Deep Knowledge** describes knowledge that has multiple dimensions, with complex relations captured within each domain. A knowledge base implements structured interactive knowledge as a repository in a particular solution, primarily implemented as knowledge graphs (e.g.: Google's Knowledge Graph [13]). Finally, an AI system with Deeply Structured Knowledge is a system with a knowledge base that captures deep knowledge and reflects its structure through extraction schemes.

The **Neural Network** (NN) is the primary functional part of a Class 3 system. It may include all perception elements such as image recognition and scene segmentation, or a language model for processing syntax, placement-based relations, and the core of common semantics. It will likely learn an embedding space that represents the key dimensions of the incoming data. In multimodal systems, it will reflect the images space and the language space. Like Class 2 systems, the neural network system in Class 3 can engage with the structured knowledge base during inference time, and retrieve the information needed to complete its task successfully. In this architecture, the training of the NN needs to be done together with the extraction mechanism and some representation of the knowledge base to allow the NN to learn how to extract the required knowledge during inference.

The **Deep Knowledge Base** contains facts and information that might be required for future inference and some or all of the deep knowledge structures depicted in Fig. 1. These include descriptive knowledge, dynamic models of the world, stories, context and source attribution, values and priorities, and Concept References. The knowledge base can change after training and can include additional data and knowledge. As long as the nature of knowledge and information is similar to that encountered by the NN during training, the modified knowledge base should be fully utilizable during inference based on its latest incarnation.

Finally, the **Reasoned Extraction** block mediates between the NN and its external source of knowledge. In the simplest case, it is a direct mapping from embedding vectors through some indexed links to the knowledge base. In the more general case, reasoned extraction will extract its information using libraries based on queries or APIs.

While promising some considerable strengths, Class 3 systems require a higher level of complexity because they necessitate creating and updating an additional element of the architecture - the deep knowledge base. They also necessitate changes to the learning algorithms because the knowledge is now split between the NN and the KB, ultimately requiring new techniques for integrating gradient descent statistical methods with symbolic representations and learning.

## 5 Thrill-K: A Blueprint for Hybrid Machine Intelligence

Advanced AI systems will integrate a full mechanism of retrieval from a large semi-structured corpus of data in addition to their knowledge base. This will require dealing with inconsistencies, incomplete knowledge, and the addition of prerequisite knowledge, based on inference processes.

From the perspective of such systems, we observe three levels of knowledge (which we call 3LK):

1. **Instantaneous knowledge** allows for rapid response to external events. This knowledge represents a direct input-to-output function that reacts to events or sequences within a well-mastered domain. All higher organisms depend on the availability of instantaneous knowledge.
2. Humans and advanced intelligent machines acquire and use **standby knowledge**. Standby knowledge requires additional processing and internal resolution within the deep knowledge base, which makes it slower than instantaneous knowledge, but applicable to a wider range of situations.
3. **Retrieved external knowledge** makes use of additional knowledge sources. Whatever the scope of knowledge is within the human brain or the boundaries of an AI system, there is substantially more information, or more recently relevant information, that can be retrieved from outside of the boundary.

### 5.1 Thrill-K's Three Levels of Knowledge

Thrill-K (contraction of "three-L-K") is a proposed architectural blueprint for AI systems that utilizes the three levels of knowledge (3LK). It provides a means for representing and accessing knowledge at three levels — in parametric memory for instantaneous knowledge, in an adjacent deeply structured knowledge base for reasoned extraction, and access to broad digital information repositories for external knowledge.

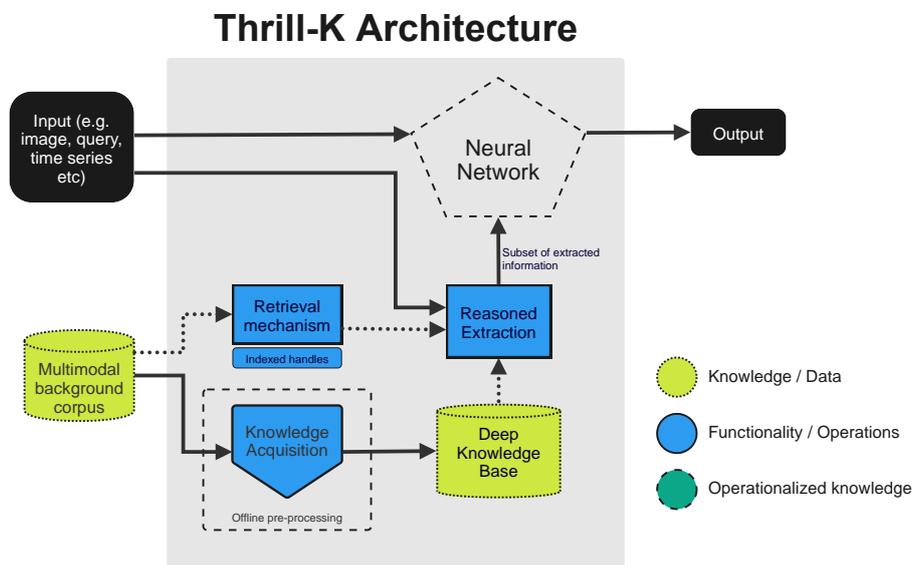

*Figure 3. Thrill-K architecture blueprint*

Fig. 3 depicts the Thrill-K system architecture. This architecture includes all the building blocks of such systems, however the flow (depicted by the arrows) can change based on the usage and configuration. In the example flow shown in the diagram, the sequence assumes the initial control is via a neural network, followed by the deep knowledge base, followed if needed by external resources. The direct input-to-output path using the instantaneous knowledge is encoded in parametric memory. If it detects uncertainty or low confidence in the direct path, the system performs reasoned extraction from its deep knowledge base. This knowledge base relies on machine-learning-based knowledge acquisition to update and refresh the knowledge as new information becomes relevant and useful enough to be added. Finally, if the AI system cannot find the knowledge needed, the retrieval mechanism enables accessing and retrieving necessary information from the available repositories. Other flows are also possible. For example, if the task of the AI is to search a knowledge base or to find paragraphs in an external repository, the same building blocks will be configured in a different sequence. It should also be noted that while the main processing path is depicted here as a neural network, the same tiering principle applies to other types of machine learning with information integrated into the processing as part of the instantaneous input-to-output path.

### 5.2    Scalability of a Thrill-K System

Aside from the obvious advantages in terms of explainability, the stratification of information access into three separate layers offers a method of mitigating compute and data costs [14] associated with scaling a typical Class 1 system. The expensive NN parametric memory is reserved only for the **instantaneous knowledge** requiring expedient access, akin to Kahneman's System 1 [4]. A Deep Knowledge base offers the necessary expansion of the scope of available **standby knowledge** at the cost of some speed of access. Finally, **retrieved external knowledge** is both the slowest (since it has to be digested through the largest number of intermediate modules), and the cheapest (since it does not have to be maintained and updated by the model itself).

By applying the 3-level knowledge hierarchy and Thrill-K system architecture, we can build systems and solutions for the future that are likely to partition knowledge at those three levels to create sustainable and viable cognitive AI.

## 6    Conclusion: Thrill-K's Contribution to Robustness, Adaptation and Higher Intelligence

While layering knowledge in three levels is essential for scale, cost, and energy, it is also required for increasing the capabilities provided by AI systems. The following are some capabilities that could be better supported by a Thrill-K system that integrates deeply structured knowledge for extraction, and access to external repositories:

1. Improved multimodal machine understanding.

2. Increased adaptability to new circumstances and tasks by retrieval/extraction of new information from repositories not available during pre-training or fine-tuning.
3. Refined handling of discrete objects, ontologies, taxonomies, causal relations, and broad memorization of facts.
4. Enhanced robustness due to the use of symbolic entities and abstracted concepts.
5. Integration of commonsense knowledge not present in the training dataset.
6. Symbolic reasoning and explainability over explicitly structured knowledge.

Thrill-K offers a new blueprint for this type of future AI architecture. It has the potential to permeate AI solutions across systems and industries and offer a method for building intelligence effectively and efficiently.

**References**

1. Devlin, J., Chang, M. W., Lee, K., & Toutanova, K. (2018). Bert: Pre-training of deep bidirectional transformers for language understanding. arXiv preprint arXiv:1810.04805.
2. Amodei, D. (2021, June 21). AI and Compute. OpenAI. https://openai.com/blog/ai-and-compute/
3. Interpreting AI compute trends. (2020, April 23). AI Impacts. https://aiimpacts.org/interpreting-ai-compute-trends/
4. Kahneman, D. (2011). Thinking, fast and slow. Macmillan.
5. Bengio, Y. (2019, December). From system 1 deep learning to system 2 deep learning. In Neural Information Processing Systems.
6. Chollet, F. (2019). On the measure of intelligence. arXiv preprint arXiv:1911.01547.
7. Launchbury, J. (2017). A DARPA perspective on artificial intelligence. Retrieved November, 11, 2019.
8. Goyal, A., & Bengio, Y. (2020). Inductive biases for deep learning of higher-level cognition. arXiv preprint arXiv:2011.15091.
9. Kautz, H. (2022). The third AI summer: AAAI Robert S. Engelmore Memorial Lecture. AI Magazine, 43(1), 93-104.
10. Lewis, P., et al (2020). Retrieval-augmented generation for knowledge-intensive NLP tasks. Advances in Neural Information Processing Systems, 33, 9459-9474.
11. Chakraborty, N., Lukovnikov, D., Maheshwari, G., Trivedi, P., Lehmann, J., & Fischer, A. (2019). Introduction to neural network based approaches for question answering over knowledge graphs. arXiv preprint arXiv:1907.09361.
12. Kapanipathi, P., et al (2020). Leveraging abstract meaning representation for knowledge base question answering. arXiv preprint arXiv:2012.01707.
13. Singhal, A. (2012). Introducing the knowledge graph: things, not strings. Official google blog, 5, 16.
14. Thompson, N. C., Greenewald, K., Lee, K., & Manso, G. F. (2020). The computational limits of deep learning. arXiv preprint arXiv:2007.05558.